\documentclass[conference]{ieeeconf}
\IEEEoverridecommandlockouts
\usepackage{cite}
\usepackage{amsmath,amssymb,amsfonts}
\usepackage{algorithmic}
\usepackage{graphicx}
\usepackage{textcomp}
\usepackage{xcolor}
\usepackage[ruled,vlined]{algorithm2e}
\usepackage{svg}
\usepackage{subcaption}
\usepackage{booktabs}
\usepackage{multirow}
\usepackage{xspace}

\def\BibTeX{{\rm B\kern-.05em{\sc i\kern-.025em b}\kern-.08em
    T\kern-.1667em\lower.7ex\hbox{E}\kern-.125emX}}
\begin{document}
\newcommand{\stSpace}{\textrm{$\mathcal{S}$}\xspace}
\newcommand{\stSpaceRes}{\textrm{${\mathcal{S}}_{r}$}\xspace}
\newcommand{\latentNSt}{\textrm{$(y, s_r)$}\xspace}
\newcommand{\latentNStP}{\textrm{$(y', s'_r)$}\xspace}

\title{POMDPs for Autonomous Science Exploration}

\author{Daniel Guirguis, Nathan Wallace, Hanna Kurniawati, and Salah Sukkarieh}
\maketitle

\begin{abstract}
Autonomous exploration missions require decision-making under sensor uncertainty and computational constraints, yet integrating scientific representations into POMDP planning has remained intractable due to high-dimensional observation spaces. Information-theoretic planners overcome this by assuming deterministic observations, sacrificing the principled uncertainty quantification that POMDPs provide. We introduce the Science Hypothesis Map POMDP (SHM-POMDP), which makes science-driven belief-space planning more tractable by branching on inferred physical properties rather than raw sensor data. This preserves full sensor information through learned observation models while enabling the planner to reason jointly about navigation and scientific properties under uncertainty. On an extended RockSample domain with 50-dimensional observations, SHM-POMDP achieves 18.6\% higher rewards and 32.9\% reduced computation time per step than continuous-observation baselines. On realistic geologic exploration using Cuprite hyperspectral data, SHM-POMDP achieves 2.5$\times$ higher information gain than the best information-theoretic baseline by maintaining beliefs and replanning adaptively---reaching 80\% of oracle performance using only uniform priors. These results demonstrate that integrating hierarchical probabilistic models into belief-space planning enables tractable, principled autonomous science that outperforms both traditional POMDP methods and science-aware information-theoretic approaches.
\end{abstract}

\section{Introduction}
Modern planetary robotic exploration relies heavily on scientists specifying waypoints based on expert knowledge and expectations about where to gather mission-critical information \cite{b1}. As measurements are collected, scientists reinterpret data with growing contextual knowledge and frequently re-plan \cite{b2}. However, in many exploration scenarios with low bandwidth and high latency communications, pre-planned waypoint navigation fails to capitalise on in-situ observations and unexpected discoveries, necessitating greater onboard autonomy \cite{b3}. 

\begin{figure}[t]
  \centering
  \includegraphics[width=\columnwidth]{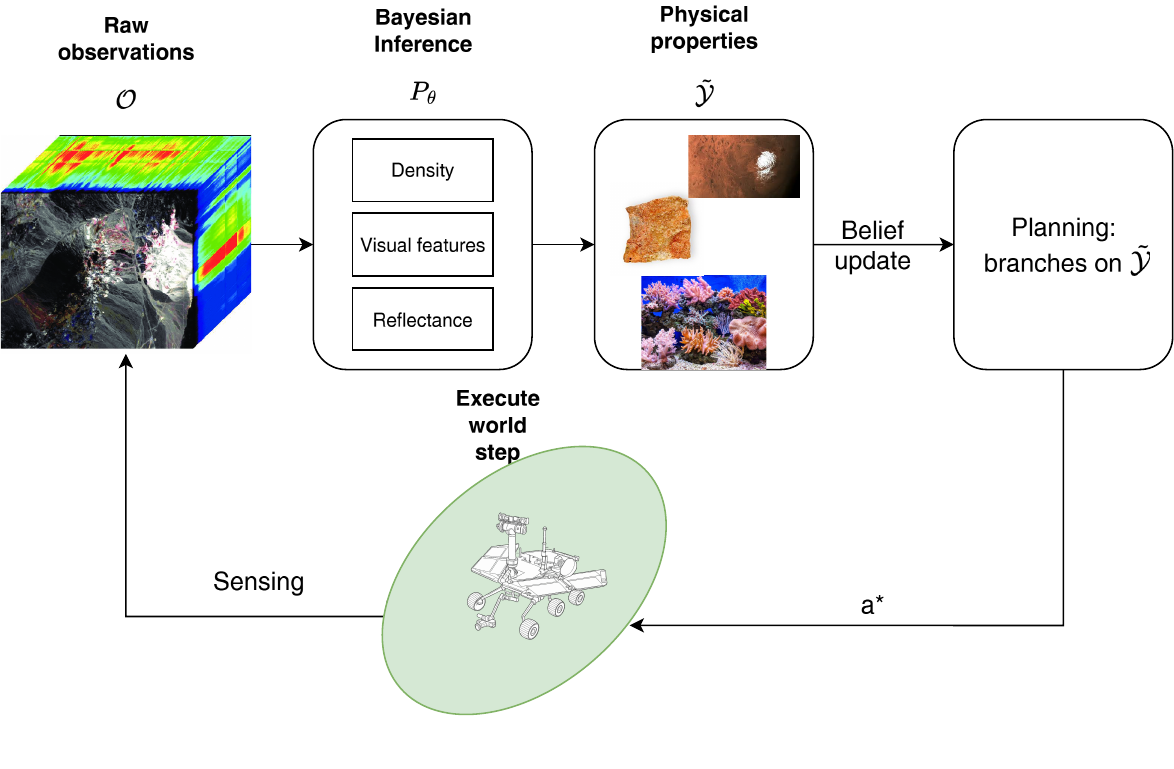}
  \caption{Overview of the SHM-POMDP framework: SHM-POMDP  makes POMDP planning tractable by planning on inferred physical properties $\tilde{y}$ rather than raw observations, abstracted through the SHM model $P_\theta$ to update the robot's belief for autonomous exploration.}
  \label{fig:flowchart}
  \vspace{-2em}
\end{figure}

This work presents a framework to overcome communication bottlenecks in robotic exploration by shifting from route-based command requiring human-in-the-loop replanning to guidance driven by an evolving model of scientific belief, enabling the robot to replan autonomously without ground-station intervention. The proposed approach accounts for uncertainty in observations, information gathering, and navigation under partial observability. The Partially Observable Markov Decision Process (POMDP) framework provides a principled approach for decision-making under uncertainty by maintaining belief distributions over states and optimising expected cumulative rewards \cite{b4}. However, solving POMDPs for scientific exploration becomes computationally challenging as multimodal sensors such as spectrometers and cameras produce high-dimensional observation vectors, and these sensors often exhibit complex correlations that further complicate direct planning. Prior work on Science Hypothesis Maps (SHM) addressed the scientific reasoning gap but assumed deterministic observations, while POMDP solvers handle uncertainty but struggle with high-dimensional observation spaces. This paper introduces the SHM-POMDP, a framework that integrates the SHM's hierarchical probabilistic structure into POMDP planning. This integration provides a principled approach for scientific exploration under observation uncertainty while improving tractability through observation abstraction, enabling autonomous exploration within the computational budget of onboard flight processors (Fig.~\ref{fig:flowchart}).

Scientists conceptualise exploration goals in terms of abstract hypotheses while robots must operate on raw sensor measurements such as spectral data, imagery, and other high-dimensional observations \cite{b5}. This gap between scientific reasoning and robotic sensing makes it difficult for autonomous systems to make decisions that are both scientifically meaningful and computationally feasible. However, it is possible to construct simpler hierarchical probabilistic models relating scientific reasoning to measurable data. For example, NASA missions formalise this relationship through the Science Traceability Matrix (STM), which employs a tripartite division into investigation objectives, physical properties, and instrument measurements \cite{b1}. For instance, an investigation objective such as ``assess past habitability'' traces to latent physical properties like organic molecule concentration, which are not directly observed but inferred from instrument measurements such as mass spectrometry readings \cite{b29}.

We exploit this hierarchical structure in SHM-POMDP by learning the mapping between robotic raw sensor measurements and latent scientific properties of interest. Standard POMDP solvers branch on raw observations, which may be high-dimensional or continuous. This leads to computational intractability as the solver creates separate branches for observations that differ only by sensor noise yet indicate the same underlying mineral, which SHM-POMDP addresses through observation abstraction, branching instead on inferred latent scientific properties. Each inferred property is a branch weighted by its posterior probability and updates use the full observation likelihood, so the belief over properties remains a distribution rather than a point estimate. This preserves the uncertainty that gives information-gathering actions their value while significantly reducing the branching complexity handled by the planner. The robot can thereby jointly reason about navigation decisions and their impact on scientific discovery while maintaining explicit uncertainty quantification throughout. We demonstrate this approach on two robotic domains: an extended rock sample problem for testing scalability, and a realistic geologic exploration task where a rover maps mineral compositions using spectral data from the Cuprite mining district. 

The contributions of this paper include: (1) extending the SHM framework to explicitly model observation uncertainty, addressing a critical limitation of prior work that assumed deterministic sensor measurements; (2) a principled integration of the SHM with POMDP planning that improves tractability through observation abstraction, branching on physical properties rather than raw sensor data; and (3) experiments validate SHM-POMDP achieving up to 18.6\% higher reward and 32.9\% reduction in computation time per step on synthetic benchmarks, and 2.5$\times$ higher information gain than the best information-theoretic baseline on realistic geologic exploration.

\section{Related work}
POMDPs provide a principled framework for decision-making under uncertainty, and Monte Carlo tree search (MCTS) methods have emerged as effective online solvers for continuous POMDPs. POMCPOW \cite{b6} handles continuous observation spaces through progressive widening, while PFT-DPW \cite{b6} uses particle filter trees for belief representation. DESPOT-$\alpha$ \cite{b7} and LABECOP \cite{b31} extend this to very large observation spaces through sparse sampling-based approaches. However, these are general-purpose solvers that operate on raw observation spaces, and when applied to robotic exploration with multimodal sensors, the observation space grows rapidly, making planning intractable. 

Hierarchical approaches address this complexity by decomposing problems into manageable subproblems, improving computational tractability \cite{b8,b9,b10,b11,b12,b14}. However,  most prior work has focused on action hierarchies, where complex behaviours are decomposed into primitives to improve tractability \cite{b12}. These methods attempt to manage observation intractability by restricting which observations each sub-POMDP processes \cite{b32}. This is insufficient because each distinct observation produces a different belief update through the POMDP likelihood model, and large observation spaces therefore increase the branching factor of the search tree---the primary driver of computational cost in online planning \cite{b15,b16}. Partitioning observations across sub-POMDPs does not resolve this, since the full observation must be maintained somewhere in the hierarchy \cite{b32}. Action-focused hierarchies therefore do not address the fundamental challenge of compressing the observation space itself.

This hierarchical structure between scientific properties and sensor measurements offers a principled basis for observation compression. While Bayesian networks have been explored to represent latent scientific properties in robotic exploration \cite{b1,b4}, they have not addressed how navigation and sensing jointly affect belief evolution. Learning-based approaches such as Differentiable Particle Filters \cite{b21} and Visual Tree Search \cite{b22} take an alternative approach, training generative observation models offline and incorporating them into online POMDP solvers to address high-dimensional observations. While effective for visual domains, these methods require large-scale training and operate in uninterpretable latent spaces, limiting their applicability to scientific missions where interpretability is essential.  

The most directly relevant work is \cite{b1}, which operationalised the STM's tripartite structure as Science Hypothesis Maps, implementing the hierarchy between investigation objectives, physical properties, and raw measurements as a Bayesian network. However, their implementation separates navigation from sensing and assumes deterministic observations, failing to capture how movement affects future observations under uncertainty. This limitation is shared by information-theoretic path planning, which provides performance guarantees but treats sensing and navigation independently \cite{b17,b18,b19}. Similar limitations are seen in deployed Mars systems such as AEGIS \cite{b13} and PIXL \cite{b20} which demonstrate autonomous capabilities but operate reactively, responding only to current observations rather than planning future actions under uncertainty. SHM-POMDP extends \cite{b1} hierarchical structure to explicitly model observation uncertainty, integrating physical properties into the POMDP observation model. This integration would be intractable without observation abstraction; by branching on physical properties rather than raw observations, we improve tractability while maintaining principled uncertainty quantification.

\section{Preliminaries}
\subsection{POMDP}
A POMDP is typically represented by the tuple ($\mathcal{S}, \mathcal{A},\mathcal {O}, T, Z, R, \gamma$), where $\mathcal{S}$ is the state space, $\mathcal{A}$ is the action space, and $\mathcal{O}$ is the observation space. The transition model $T(s, s', a) = P(s' \mid s, a)$ specifies the conditional probability of moving from state $s$ to $s'$ after performing action $a$, while the observation model $Z(s', a, o) = P(o\mid s',a)$ specifies the conditional probability of observing $o$ after taking action $a$ in state $s'$. The reward function $R(s, a)$ quantifies the utility of executing action $a$ in state $s$. The discount factor $\gamma \in [0,1)$ balances immediate and future rewards. The agent maintains a belief distribution $b_t(s)$ over states, updated after each action-observation pair through Bayes' rule \cite{b23}:
\begin{equation}
    b_{t+1}(s') \propto Z(s', a_t, o_t) \sum_{s \in \mathcal{S}}T(s, s', a_t)b_t(s).
\end{equation}

\subsection{Science Hypothesis Map}
Scientific exploration requires reasoning about physical properties that cannot be measured directly but must be inferred from sensor data. We formalise this relationship through a SHM, adapting the hierarchical probabilistic structure introduced by \cite{b1} for integration with online POMDP planning.

The SHM defines a three-level hierarchy  $\mathcal{H} \rightarrow \mathcal{Y} \rightarrow \mathcal{O}$ relating scientific concepts to observable measurements. At the highest level, investigation objectives $\mathcal H = \{h_1, \ldots, h_l \}$ represent the scientific hypotheses a mission seeks to evaluate---for instance, determining whether a region formed through hydrothermal or sedimentary processes. At the intermediate level, latent physical scientific properties $\mathcal Y = \{y_1, \ldots, y_m \}$ are attributes that serve as diagnostic indicators of these objectives; in geologic exploration, these are mineral types such as chlorite, serpentine or calcite. These properties are physical (grounded in measurable reality), scientific (meaningful to domain experts), and latent (partially observed through noisy sensors). For brevity, we refer to these as physical properties throughout. At the lowest level, the observation space $\mathcal{O}$ captures raw sensor observations such as hyperspectral reflectance data, where $\mathcal{O} \subseteq \mathbb{R}^d$ and $d$ is the number of spectral bands.

These three levels are linked by conditional distributions. The distribution $P(y\mid h)$ specifies how physical properties relate to investigation objectives, such as which minerals are characteristic of which geologic formations. The observation likelihood $P_\theta(o \mid y)$, parameterised by learned parameters $\theta$, specifies the distribution over sensor measurements given the underlying physical property. Together, these define a joint distribution that factorises as:
\begin{equation}
    P(h,y,o) = P(h)\cdot P(y\mid h) \cdot P_\theta(o \mid y).
\end{equation}
While the full hierarchy supports inference up to investigation objectives, the planning algorithm we present operates at the $\mathcal{Y} \rightarrow \mathcal{O}$ interface, using physical properties as the abstraction level for observation branching. This level provides sufficient abstraction for computational tractability while remaining grounded in scientifically meaningful categories.

The physical property space $\mathcal{Y}$ plays a central role in our formulation, serving as the bridge between raw sensor data and scientific interpretation. A physical property $y \in \mathcal{Y}$ is an attribute of the environment that satisfies three criteria. First, it must be observationally grounded, meaning distinct properties induce distinct measurement distributions: $P_\theta(o\mid y_i) \neq P_\theta(o\mid y_j)$ for any $y_i \neq y_j$. This guarantees that the property can be inferred from sensor data. Second, it must be interpretable, corresponding to a discrete set of domain-meaningful concepts $\mathcal{C}$ (e.g., mineral types, chemical compositions, or terrain classes) such that $\mathcal{Y} \subseteq \mathcal{C}$, rather than an arbitrary continuous latent variable. Third, in the context of exploration missions, a physical property is static: the robot's actions do not alter the underlying property, which is formally defined by the transition deterministic mapping $T(y, a, y') = 1$ if $y = y'$ and $0$ otherwise.

\section{SHM-POMDP}
\begin{algorithm}[t]
\small
\caption{SHM-POMDP Planning and Execution}
\label{alg:shm-pomdp-revised}
\KwIn{Belief $b$, depth $d$, SHM model $P_\theta(o|y,s',a)$, threshold $\epsilon$, simulations $N$}
\KwOut{Action $a^*$ and updated belief $b'$}
\SetKwProg{Fn}{Procedure}{:}{}
\Fn{TreeSearch($b,d$)}{
  \If{$d = 0$ or IsTerminal($b$)}{
    \KwRet{0}
  }
  \If{$b \notin$ Tree}{
    InitialiseNode($b$)\\
  }
  $a \gets$ SelectAction($b$)\\
  $s = (y, s_r) \sim b$\\
  $b' \gets$ PredictBelief($b, a$)\\
  $s' = (y', s'_r) \sim b'$\\
  $o_{exp} \gets$ ComputeExpectedObs$(s')$\\
  \ForEach{$\tilde{y}_i \in \tilde{\mathcal{Y}}$}{
    $p_i \gets P_\theta(o_{exp}|\tilde{y}_i, s', a) \cdot P(\tilde{y}_i|s', a)$\\
  }
  Normalise$(p_1, \ldots, p_k)$ to get $P(\tilde{y}|o_{exp}, s',a)$\\
  $\tilde{\mathcal{Y}}_{branch} \gets \{\tilde{y}_i : P(\tilde{y}_i|o_{exp}, s',a) > \epsilon\}$\\
  $Q \gets R(s,a)$\\
  \ForEach{$\tilde{y}_i \in \tilde{\mathcal{Y}}_{branch}$}{
    \If{$(a,\tilde{y}_i) \notin$ Children($b$)}{
      $b_{child} \gets$ UpdateBelief($b', a, \tilde{y}_i$)\\
      AddChild($b, a, \tilde{y}_i, b_{child}$)\\
    }
    \Else{
      $b_{child} \gets$ GetChild($b, a, \tilde{y}_i$)\\
    }
    $Q \gets Q + \gamma \cdot P(\tilde{y}_i|o_{exp}, s',a) \cdot$ TreeSearch($b_{child}, d-1$)\\
  }
  $Q(b,a) \gets Q$\\
  \KwRet{$Q(b,a)$}
}
\Fn{Execute($b, a^*$)}{
  Execute $a^*$ in environment\\
  $o \gets$ GetObservation()\\
  \ForEach{particle $s_i = (y_i, s_{r,i})$ in $b$}{
    $s'_i = (y'_i, s'_{r,i}) \sim T(s'|s_i, a^*)$\\
    $w'_i \gets w_i \cdot P_\theta(o \mid y'_i, s'_i, a^*)$\\
  }
  \KwRet{WeightedResample$(\{s'_i, w'_i\})$}
}
\Repeat{mission complete}{
  \tcp{plan}
  \For{$n \gets 1$ \KwTo $N$}{
    TreeSearch($b, d$)\\
  }
  \tcp{execute}
  $b \gets$ Execute$(b, \arg\max_a \, Q(b, a)$)\\
}
\end{algorithm}
We integrate the Science Hypothesis Map with the POMDP to enable science-driven planning under uncertainty; we term the resulting formulation an SHM-POMDP. The state space factors as $\stSpace = \mathcal{Y} \times \stSpaceRes$, where $\mathcal{Y}$ represents physical properties and \stSpaceRes is the remaining state containing all other task-relevant variables. A complete state $s = (y, s_r)$ specifies both the physical property of interest and the robot's operational context. The variables $y \in \mathcal{Y}$ are partially observed, while $s_r \in \stSpaceRes$ may be fully observed, partially observed, or a mixture of both, depending on the application.

The transition function is then defined as 
$T(\latentNSt, a, \latentNStP) = T(y, a, y') \times T(s_r, a, s'_r)$, where $T(y, a, y') = 1$ for $y = y'$ and $0$ otherwise, as physical properties remain static during the mission.

The SHM defines a generative model $P_\theta(o \mid y)$ specifying how physical properties produce observations. Within the SHM-POMDP, this model is extended to $P_\theta(o \mid y, s', a)$ to account for measurement conditions such as distance to target and sensor noise. During planning, the robot predicts the belief forward through the transition model (PredictBelief), then inverts the observation model to reweight the predicted belief into a posterior over property values (UpdateBelief), applying no further transition. Here PredictBelief gives $b'(s') = \sum_s T(s' \mid s,a)b(s)$ and UpdateBelief reweights, $b_{\text{child},i}(s') \propto P(\tilde{y_i} \mid s',a)b'(s')$ then resamples. We denote these inferred values as $\tilde{y} \in \tilde{\mathcal{Y}}$ to distinguish the robot's inferred estimate from the true physical property $y$ in the state. The inferred set $\tilde{\mathcal{Y}}$ need not equal $\mathcal{Y}$---it may refine $\mathcal{Y}$ or add an uncertain outcome---but each element is scientifically meaningful and $\mid\tilde{\mathcal{Y}}\mid \ll \mid \mathcal{O}\mid$.

To formalise this inversion, we augment the standard POMDP observation function to include the inferred property as an intermediate variable. The observation function becomes:
\begin{equation}
    Z(s', a, o) = P(o,\tilde{y} |s',a) = P_\theta(o \mid \tilde{y},s',a) \cdot P(\tilde{y} \mid s',a), 
\end{equation}
where $P_\theta(o \mid \tilde{y},s',a)$ is the SHM's likelihood of observations given the inferred property and measurement conditions, and $P(\tilde{y} \mid s',a)$ is the prior distribution over inferred properties given the sampled state. This factorisation allows us to compute the posterior distribution over inferred properties via Bayes' rule. For a state $s' = (y',s_r')$ drawn from the predicted belief $b'$, an expected observation $o_{\text{exp}}$ (ComputeExpectedObs) is generated. The posterior over inferred properties given this expected observation is:
\begin{equation}
    P(\tilde{y}_i\mid o_\text{exp}, s',a) = \frac{P_\theta(o_\text{exp}\mid \tilde{y}_i,s',a)\cdot P(\tilde{y}_i\mid s',a)}{\sum_j P_\theta(o_\text{exp}\mid \tilde{y}_j,s',a)\cdot P(\tilde{y}_j\mid s',a)},
\end{equation}
where $\tilde{y}_i$ is a candidate inferred property, $o_\text{exp}$ is the characteristic observation from the sampled state $s'$, $P_\theta$ is the learned SHM generative observation model, $P(\tilde{y}_i \mid s',a)$ is the prior probability of observing $\tilde{y}_i$ given the sampled state and action, and $j$ indexes over all possible properties in $\tilde{\mathcal{Y}}$.

We use the computed distribution $P(\tilde{y}_i \mid o_\text{exp},s',a)$ to determine which inferred properties to branch on. The algorithm identifies all $\tilde{y}_i$ values with non-negligible probability, specifically $\{\tilde{y}_i: P(\tilde{y}_i \mid o_{\text{exp}},s',a) > \epsilon \}$ where $\epsilon$ is a threshold parameter. The solver then creates branches for each of these inferred properties, treating them as discrete outcomes rather than raw sensor data \cite{b6}. In the resulting tree structure, nodes represent belief states $b$ over $(y,s_r)$, and edges represent action-observation pairs $(a,\tilde{y}_i)$ where $\tilde{y}_i$ is an inferred property. Each branch ($a,\tilde{y}_i$) leads to a child belief representing the posterior after inferring property $\tilde{y}_i$. The tree thus encodes uncertainty over physical properties that might be inferred, rather than specific sensor readings. This abstraction reduces the branching factor from the size of the continuous observation space $|\mathcal{O}|$ to the number of likely inferred properties $\lvert\{\tilde{y}_i: P(\tilde{y}_i \mid o_{\text{exp}},s',a) > \epsilon \}\rvert \leq |\tilde{\mathcal{Y}}|$, where $|\tilde{\mathcal{Y}}| \ll |\mathcal{O}|$. Algorithm \ref{alg:shm-pomdp-revised} provides the complete procedure for SHM-POMDP tree search with observation abstraction.

The probability $P(\tilde{y}_i\mid o_\text{exp},s',a)$ serves as a weight that determines which branches are created and how much each branch contributes to the expected value calculation during planning. While the use of a single representative observation $o_\text{exp}$ to determine branches is an approximation during planning, the full observation likelihood $P_\theta(o \mid y,s',a)$ is used without approximation during execution for belief updates, preserving sensing fidelity where it most affects downstream planning. During tree search, the value function aggregates over branches as $Q(b,a) = R(s,a) + \gamma\sum_i P(\tilde{y}_i \mid o_\text{exp}, s',a)  \cdot V(b_\text{child,i})$, where $s \sim b$ is the state sampled at this node and each child belief represents the posterior after discovering $\tilde{y}_i$. This weighted aggregation ensures the planning process accounts for the full distribution of possible inferences while achieving computational efficiency through discrete branching.

During execution, the robot takes the best action $a^*$ determined by the tree search and receives a high-dimensional observation $o \in \mathbb{R}^d$ from its sensors. Rather than using the inferred properties from planning, the belief update employs the full observation likelihood from the SHM. Particle weights are updated as $w'_i = w_i \cdot P_\theta(o\mid y_i',s'_{i},a^*)$ using the SHM likelihood, then resampled to concentrate on likely classifications. This combination of discrete branching for planning and full observations for execution achieves tractability without sacrificing sensing fidelity.

The SHM provides a learnable probabilistic generative observation model that can be learned offline from training data consisting of paired observations and ground-truth labels. The model captures the conditional distribution $P_\theta(o\mid y,s',a)$, which represents the likelihood of observing data $o$ given a physical property $y$, measurement conditions $s'$ and action $a$. In the experimental domains presented, all sensing actions use identical observation mechanisms, $P_\theta(o\mid y,s',a) = P_\theta(o\mid y,s')$ for all $a$. The specific implementation can vary by domain---Gaussian Mixture Models for spectral data, neural networks for images, or other probabilistic models appropriate for the sensor modality.

\section{Experiments}
We evaluate SHM-POMDP across two domains that address the challenge of autonomous exploration in high-dimensional observation spaces where scientific reasoning is required. The robot must infer physical properties from noisy, continuous high-dimensional sensor data while jointly optimising navigation and information gathering under budget constraints. Since movement incurs a step penalty, cumulative reward serves as a joint proxy for scientific and navigational efficiency, where higher reward reflects effective and efficient paths to achieve them. Section \ref{sec:rocksamplepomdp} uses an extended RockSample benchmark with continuous multi-modal observations to demonstrate scalability: as observation dimensions grow from 5D to 50D, SHM-POMDP maintains constant branching factor $O(\mid \tilde{\mathcal{Y}} \mid^d)$ while standard solvers face $O(\mid \mathcal{O} \mid^d)$. Section \ref{sec:Cupriteexploration} uses realistic geologic mapping with authentic spectral confusion to compare SHM-POMDP against state of the art information-theoretic planners; when spectral overlap creates ambiguity, belief-based planning detours to disambiguate while static planners cannot adapt.

We assess two aspects of performance. First, computational efficiency captures planning time per decision step, reflecting feasibility for onboard processors with limited computational budgets. Second, scientific return measures the value of exploration: cumulative reward for RockSample and realised information gain for Cuprite. Statistical significance is assessed via paired t-tests ($\alpha = 0.05$) across 100 trials with identical random seeds. All experiments were conducted on a laptop with an Intel Ultra 7 Processor 155H 4.8 GHz and 32GB RAM.

\subsection{Extended RockSample}
\label{sec:rocksamplepomdp}

\subsubsection{Setup}
\label{sec:firstpomdp}
We modify the Information Search RockSample (ISRS) domain \cite{b29} to model a rover exploring an $n \times n$ grid with $k$ rocks of unknown scientific value (good or bad). The state $s = (y, s_r)$ consists of latent rock types $y \in \{good, bad, uncertain\}^k$ (partially observable) and $s_r = (x_\text{rover},y_\text{rover}, \{(x_i,y_i)\}^k_{i=1})$, which contains the rover position and rock locations (fully observable). The rover can move in the four cardinal directions, sample rocks for reward, and sense rocks with continuous-valued observations whose fidelity decays with distance. Movement incurs a penalty, requiring efficient path planning that balances information gathering against traversal costs.

We simulate realistic sensing through five modalities with distinct noise characteristics and structured intra-modality correlations. The rover's multi-sensor observation is modelled as a multivariate normal whose mean encodes the expected rock signature and whose covariance captures distance dependent noise \cite{b23}. While rock states are binary, the SHM-POMDP maps these high-dimensional observations to three abstracted values $\tilde{y} \in \{good, bad, uncertain\}$ for tree branching. High confidence observations map to $\tilde{y} = good \text{ or } bad$, while ambiguous observations yield $\tilde{y} = uncertain$. This 3-way branching reduces computational complexity while preserving full observational information $P_\theta(o\mid y,s')$ for belief updates.

We compare SHM-POMDP against a continuous-observation POMDP baseline, both using the POMCPOW solver. The key difference is that SHM-POMDP branches on abstracted observations $\tilde{y}$ while the baseline branches on continuous observations $o$. Hyperparameters including $\epsilon$, were tuned on 5D observations via preliminary sweeps and fixed for higher dimensions to isolate scaling effects, yielding $\epsilon = 10^{-4}$. Both methods used identical action selection, depth limits, and particle counts to ensure comparable computational requirements.

\subsubsection{Results}

\begin{table}[t]
\centering
\caption{Scalability comparison: SHM-POMDP vs Continuous baseline across increasingly difficult RockSample problems (mean $\pm$ 95\% CI, 100 trials). SHM-POMDP achieves up to 18.6\% higher reward and 32.9\% reduction in computation time per step as observation dimensionality increases.}
\label{tab:results}
\resizebox{\columnwidth}{!}{%
\begin{tabular}{@{}lllcccc@{}}
\toprule
& & & \multicolumn{2}{c}{\textbf{Reward}} & \multicolumn{2}{c}{\textbf{Time/step (sec/step)}} \\
\cmidrule(lr){4-5} \cmidrule(lr){6-7}
\textbf{Env} & \textbf{Rocks} & \textbf{Dim} & CONT & SHM & CONT & SHM \\
\midrule
\multirow{4}{*}{5×5}
& \multirow{4}{*}{4}
& 5D  & 8.58±0.60 & \textbf{9.04±0.73} & \textbf{0.22} & 0.23 \\
& & 10D & 8.08±0.75 & \textbf{8.99±0.74} & 0.38 & \textbf{0.31} \\
& & 30D & 8.59±0.72 & \textbf{9.81±0.82} & 1.00 & \textbf{0.82} \\
& & 50D & 8.27±0.85 & \textbf{8.86±0.87} & 1.26 & \textbf{0.90} \\
\midrule
\multirow{4}{*}{7×7}
& \multirow{4}{*}{8}
& 5D  & 6.66±0.92 & \textbf{7.34±0.82} & 0.45 & 0.45 \\
& & 10D & 8.51±1.05 & \textbf{8.85±1.01} & 0.77 & \textbf{0.63} \\
& & 30D & 8.47±0.89 & \textbf{8.99±0.96} & 2.18 & \textbf{1.67} \\
& & 50D & 7.19±0.88 & \textbf{7.35±0.92} & 2.57 & \textbf{1.69} \\
\midrule
\multirow{4}{*}{15×15}
& \multirow{4}{*}{15}
& 5D  & \textbf{3.13±1.61} & 3.09±1.62 & 0.60 & \textbf{0.59} \\
& & 10D & 3.51±1.68 & \textbf{3.92±1.58} & 1.01 & \textbf{0.73} \\
& & 30D & 3.56±1.65 & \textbf{4.07±1.56} & 3.52 & \textbf{2.72} \\
& & 50D & 2.79±1.57 & \textbf{3.31±1.62} & 6.81 & \textbf{4.57} \\
\bottomrule
\end{tabular}%
}
\vspace{-2em}
\end{table}
Table \ref{tab:results} summarises performance across systematically scaled problem complexity ($5\times 5, 7 \times 7, 15 \times 15 \text{ maps})$ and observation dimensionality (5D - 50D). Both methods executed identical episode counts per planning step using 100 identical random seeds.

SHM-POMDP's primary advantage emerges in computational efficiency, with gains that scale dramatically with problem complexity and observation dimensionality. The computational efficiency showed minimal advantages on small problems ($5 \times 5$ with 5D: -4.5\%) but substantial improvements on complex problems ($15 \times 15$ with 50D: 32.9\%). This scaling reflects the different architecture of the algorithms as standard POMCPOW's progressive widening produces highly varied samples in high-dimensional spaces, forcing the tree wide and shallow by fragmenting computation across $O(|\mathcal{O}|^d)$ branches where $d$ is the search depth. SHM-POMDP circumvents this explosion through hierarchical decomposition. During tree expansion, the planner branches only over the abstracted observations $\tilde{\mathcal{Y}}$, maintaining constant $O(|\tilde{\mathcal{Y}}|^d) = O(3^d)$ branching regardless of observation dimensionality. When an observation is received during rollout, SHM-POMDP evaluates the likelihood $P_\theta(o\mid y,s')$ through the pre-defined sensor model without creating explicit tree branches for each possible observation. This architectural difference grows on complex $15 \times 15$ maps with many rocks and long horizons, where deeper search matters most.

The observed reward improvements (up to 18.6\% in $15 \times 15$, 50D) reflect more efficient navigation to valuable rocks, as reduced branching enables deeper tree search under fixed computational budget, revealing path value differences shallow trees cannot distinguish. Additionally, SHM-POMDP exploits sensor correlations encoded in the observation model. The observation model $P(o\mid y,s')$ incorporates both intra-modal (specified correlation coefficients within each sensor type) and inter-modal (all sensors respond proportionally to the same underlying rock property) correlations. This correlation structure means numerically different but physically equivalent observations naturally cluster into three abstracted values $\tilde{y}$, redirecting computation from redundant distinctions towards deeper search. 

The reward improvements are largest in the 10--30D range rather than at 50D. At intermediate dimensions, progressive widening creates enough branches to fragment computational effort across observation distinctions, yet not enough to cover the space meaningfully---the worst case for unstructured observation handling, which SHM-POMDP's fixed branching circumvents. At 50D, the gap narrows not because SHM-POMDP degrades, but because the solver's behaviour changes. With extreme sparsity, every sampled observation is unique, so progressive widening creates a very wide tree that restricts search depth. SHM-POMDP's smaller branching factor still enables a deeper search, as reflected in the computational efficiency gain.

\section{Cuprite Geologic Exploration Path Planning}
\label{sec:Cupriteexploration}

\subsubsection{Setup}
\label{sec:experimental}
We evaluate our approach using the Cuprite Hills mining district, Nevada; a well characterised site with extensive ground-truth mineralogical maps and airborne hyperspectral imagery. The site contains diverse hydrothermal alteration zones containing distinctive mineral assemblages, providing a challenging yet realistic testbed for autonomous exploration \cite{b26}. We discretise a $1\times 1$\,km region into a $10\times10$ grid where each cell represents $100 \times 100$\,m of terrain.

The POMDP state comprises mineral types $y$ for each of 10 rocks scattered across the map (partially observable) and $s_r$ containing the rover and rock locations (fully observable). The rover must identify mineral classes of interest (e.g., alunite, chlorite) from hyperspectral reflectance vectors $o$ spanning the visible to short-wave infrared range. Many minerals exhibit characteristic absorption features due to their molecular composition, enabling supervised learning to associate spectra with mineral labels. 

To construct the SHM observation model, we aligned USGS mineralogical maps \cite{b1} with co-registered Next Generation Airborne Visible Infrared Imaging Spectrometer (AVIRIS-NG) hyperspectral data \cite{b27}, yielding pixel-level $(y, o)$ pairs across five mineral classes. We applied Principal Component Analysis (PCA) to reduce dimensionality while preserving diagnostic variance, then fit a Gaussian Mixture Model (GMM) with one component per class to obtain $P_\theta(o \mid y)$. This captures both the central tendency of each mineral's spectral signature and the covariance structure induced by sensor noise and natural variability \cite{b28}. During online planning, we augment this model with distance-dependent observation noise to create $P_\theta(o\mid y,s')$, where measurement quality degrades with rover-to-rock distance. The branching threshold $\epsilon = 10^{-6}$ was selected via preliminary tuning to account for the finer spectral distinctions between mineral classes. This formulation provides likelihoods that can be used for Bayesian belief updates in the POMDP. It also accounts for inter-class overlap by representing minerals as overlapping Gaussian components rather than hard classification boundaries.

\begin{figure}[t]
  \centering
  \includegraphics[width=0.65\columnwidth]{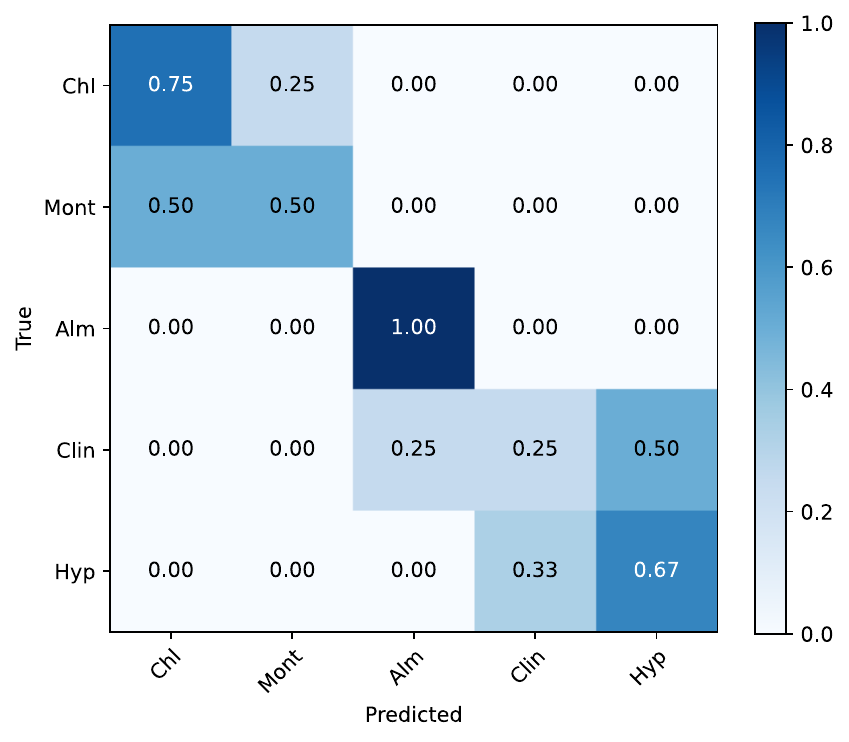}
  \caption{Confusion matrix for five-mineral LDA classifier trained on AVIRIS-NG spectra. Off-diagonal values indicate spectral overlap between minerals (e.g., Chlorite–Montmorillonite), demonstrating the observation uncertainty that SHM-POMDP explicitly models.}
  \label{fig:mineral-confusion}
  \vspace{-2em}
\end{figure}

To validate the spectral classification, we trained a linear discriminant analysis (LDA) classifier on five representative minerals using AVIRIS-NG. The confusion matrix in Fig.~\ref{fig:mineral-confusion} illustrates classification errors due to overlapping absorption features. These results underscore the need for a probabilistic observation model, as a hard classifier would obscure uncertainty in spectral interpretation \cite{b30}. Similar to the VTS framework \cite{b22}, we combine offline-learned probabilistic models with online POMDP planning, but here our SHM enables the planner to branch on abstracted observations while still maintaining beliefs over true mineral states.

In each trial, 10 rocks are randomly placed on the discretised Cuprite Hills grid. The rover navigates from the northwest to the southeast corner under a budget of twice the start to goal Manhattan distance, with the objective of maximising information gain about chlorite presence. The rover receives a terminal reward for reaching the goal and a per-step penalty so that information gathering is traded against reaching the goal within budget. The true mineral state $y \in \{\text{chlorite, not-chlorite}\}$ is the target variable, while abstracted observations $\tilde{y} \in $ \{chlorite (chl), montmorillonite (mont), almandine (alm), clinochlore (clin), hypersthene (hyp)\} represent classification outcomes used for tree branching. Information gain is computed as:
\begin{equation}
    IG = H(y_\text{chlorite}) - H(y_\text{chlorite}\mid o).
\end{equation}
Where $H$ denotes binary entropy in nats and posteriors are updated using the learned GMM observation model $P_\theta(o\mid y,s')$. Per-trial information gain is computed as the cumulative sum of per-step values across all sensing actions in the episode, where entropy is evaluated using the particle belief approximation maintained by the planner.

We do not repeat the POMDP comparison from Section \ref{sec:rocksamplepomdp} as the computational advantages remains unchanged. Instead, we compare against science-aware information-theoretic path planners to test SHM-POMDP's ability to handle real spectral confusion through adaptive belief-based planning versus static information-gain heuristics. We evaluate two scenarios: (1) uniform prior, where all planners begin with $p(\text{chlorite}) = 0.2$, testing performance under maximum uncertainty; and (2) oracle initialisation, where baseline planners (Greedy, pSPIEL) receive ground-truth mineral maps for planning, representing best-case performance in a perfect information `oracle planning' scenario. The oracle baselines receive ground-truth maps as a deliberate best-case upper bound. They compute optimal paths assuming perfect knowledge but still receive noisy observations during execution. Notably, SHM-POMDP uses only uniform priors in both cases. The fundamental difference is that baseline planners follow pre-computed paths regardless of observations, while SHM-POMDP updates beliefs and replans based on actual spectral measurements. We compare three science-aware planners that incorporate mineralogical information into their decision making, and one science-blind baseline. \textit{Random} selects actions uniformly at random. \textit{pSPIEL} uses submodular orienteering approach based on \cite{b17}, with budget-aware lookahead. \textit{Greedy} selects neighbouring cells maximising expected entropy reduction. \textit{SHM-POMDP} maintains full belief distribution and replans under uncertainty.

\subsubsection{Results}
\begin{table}[t]
\centering
\caption{Performance comparison on Cuprite geologic exploration (mean $\pm$ 95\% CI, 100 trials). Realised information gain is measured in nats; path length is the number of grid cells traversed. SHM-POMDP with uniform prior achieves 2.5$\times$ higher information gain than the best baseline and 80\% of oracle performance.}
\label{tab:planner-results}
\begin{tabular}{lcc}
\toprule
Planner & Realised IG (nats) & Path length (cells) \\
\midrule
\multicolumn{3}{l}{\textit{Uniform prior}} \\
Greedy             & $0.242 \pm 0.053$ & $19.1 \pm 0.1$ \\
Random             & $0.156 \pm 0.061$ & $30.8 \pm 1.1$ \\
pSPIEL             & $0.305 \pm 0.080$ & $30.3 \pm 0.6$ \\
SHM-POMDP          & $\mathbf{0.760 \pm 0.156}$ & $31.8 \pm 4.9$ \\
\midrule
\multicolumn{3}{l}{\textit{Ground truth prior}} \\
Greedy (oracle)    & $0.935 \pm 0.097$ & $31.2 \pm 0.5$ \\
pSPIEL (oracle)    & $0.944 \pm 0.101$ & $30.8 \pm 0.5$ \\
\bottomrule
\end{tabular}
\vspace{-2em}
\end{table}
Table \ref{tab:planner-results} presents the performance comparison across all trials. SHM-POMDP achieves significantly higher information gain ($0.760 \pm 0.156$ nats) compared to the best baseline planner pSPIEL ($0.305 \pm 0.080$ nats), representing a $2.5\times$ improvement while maintaining comparable path lengths.

SHM-POMDP's advantage stems from its ability to maintain and update beliefs during execution. Heuristic planners collapse each measurement to a single label via maximum likelihood, discarding measurement confidence levels and treating ambiguous spectra as certain. When noise or spectral overlap causes deviations from expectation, these planners cannot adapt, causing overconfident early choices that cause errors to propagate into the rest of the trajectory.  

Notably, SHM-POMDP with only uniform priors achieves 80\% of the information gain obtained by oracle planners (0.760 vs 0.944 nats). This performance gap reveals the fundamental limitation of static planning: even with perfect prior knowledge, static planners cannot adapt when observations deviate from expectations. SHM-POMDP maintains beliefs through particle filtering weighted by $P_\theta(o\mid y,s')$, where each observation reweighs particles and drives exploration toward observations that disambiguate between chlorite and non-chlorite. Particle diversity encourages exploration in uncertain regions, and noisy observations are naturally integrated across the belief distribution, producing robustness absent in heuristic methods. Because SHM-POMDP computes a policy rather than a fixed path, its behaviour adapts to realised observations, explaining the higher variance in path length compared to deterministic baselines.

\begin{figure}[t]
  \centering
  \includegraphics[width=0.65\columnwidth]{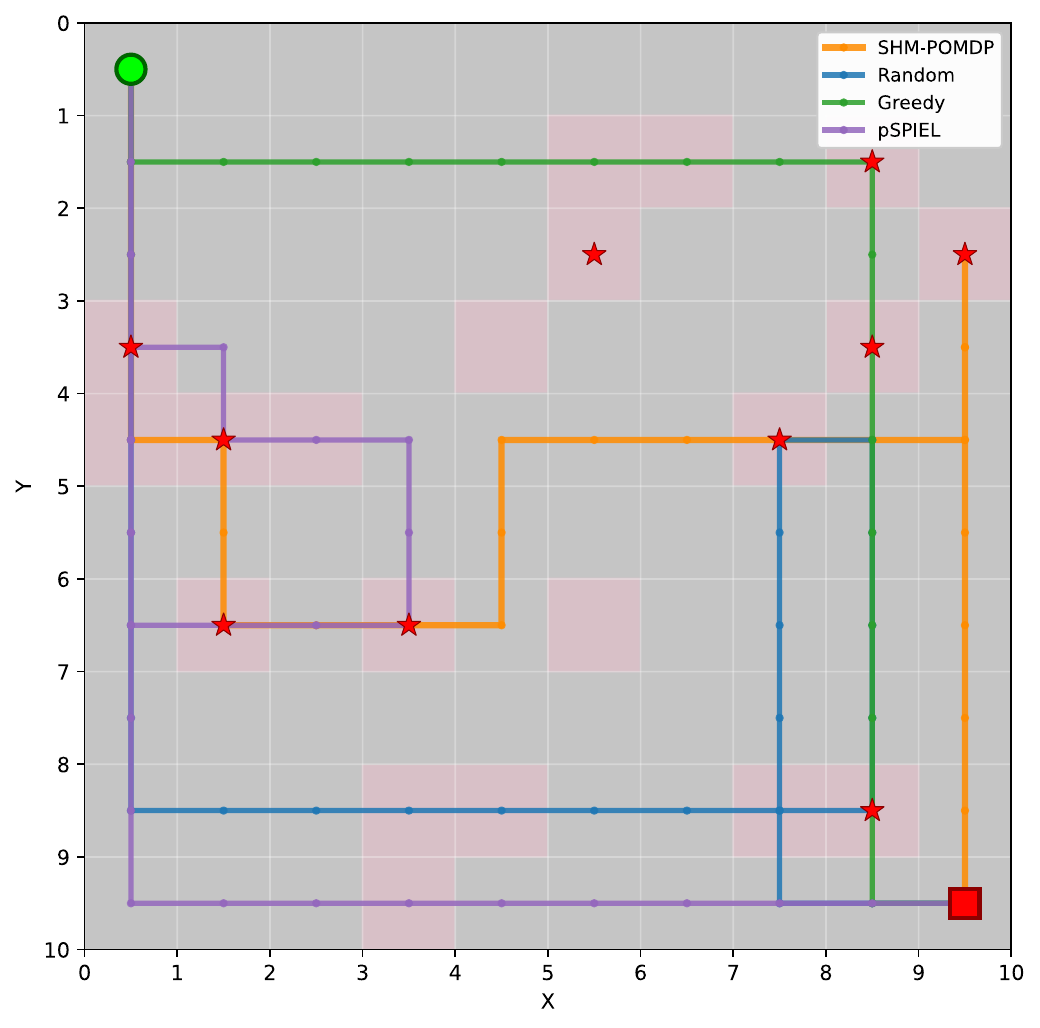}
  \caption{Cuprite exploration paths (start: north-west; goal: south-east). Belief-based planning (SHM-POMDP) enables deliberate detours to disambiguate uncertain observations, unlike static information-theoretic planners.}
  \label{fig:paths}
  \vspace{-2em}
\end{figure}

Fig.~\ref{fig:paths} illustrates how algorithmic differences manifest in path structure. With only a uniform prior, heuristic planners cannot discriminate between rocks before observing them, so their strategies reflect static heuristics rather than adaptive reasoning. Random achieves 3/10 detections purely through chance. Greedy's short trajectory (19 steps, 3/10 detections) reflects myopic selection by IG(rock) - $\lambda \times \text{detour cost}$, leaving many rocks unexplored. pSPIEL achieves only 4/10 detections despite its submodular optimisation. It groups rocks into clusters and plans systematic coverage within each cluster, but the uniform prior provides no guidance about which clusters actually contain valuable minerals. This produces consistent short paths with low variance (around 30 steps on average), but much of this efficiency is spent covering a singular uninformative region while missing high-value areas identifiable through adaptive belief updates, since its precomputed plan cannot respond when observations deviate from expectation.

In contrast, SHM-POMDP produces longer and more variable paths (around 32 steps with higher variance) that emerge from belief updates during execution. Starting from the same uninformative prior, the planner gathers initial observations to identify promising regions, then systematically explores areas where updated beliefs indicate chlorite probability exceeds the baseline. The difference is not path shape but the reason for revisiting. Random revisits by chance; pSPIEL revisits due to precomputed cluster coverage; SHM-POMDP revisits in response to observations. When an ambiguous measurement increases uncertainty, the planner deliberately seeks additional observations to disambiguate, building an increasingly accurate belief map. This adaptive replanning explains both the additional path length and the significant information gain advantage.

\section{Conclusion}
SHM-POMDP enables autonomous scientific exploration by integrating domain knowledge into belief-space planning through hierarchical probabilistic models. First, the SHM's physical property space provides a principled basis for observation abstraction, enabling branching on physical properties rather than raw sensor data to avoid the computational explosion that cripples traditional POMDP solvers while preserving full sensor information through learned observation models that capture both inter-modal and intra-modal correlations. Second, explicit modelling of observation uncertainty enables adaptive exploration that myopic and heuristic methods cannot achieve. The ability to maintain full belief distributions rather than point estimates allows the planner to recognise the value of partial information and adapt strategies based on realised observations.

Future work should extend SHM-POMDP beyond simple categorical physical properties to richer representations with probabilistic dependencies, thereby better capturing complex relationships between multi-modal observation and underlying phenomena. Online adaptation of observation models would increase robustness when sensor characteristics drift or new environments differ from training data. As missions venture to increasingly remote environments where communication latency precludes human intervention, frameworks enabling principled reasoning about scientific uncertainty will become essential for maximising discovery potential.

\bibliographystyle{IEEEtran}
\bibliography{refs}

\vspace{12pt}

\end{document}